\documentclass[journal]{IEEEtran}
\ifCLASSINFOpdf
\else
\fi
\usepackage{multirow}
\usepackage{graphicx}
\usepackage{subfigure}
\usepackage{amsmath}
\usepackage{mathrsfs}
\usepackage{CJK}
\usepackage{epsfig}
\usepackage{amsmath}
\usepackage{amssymb}
\usepackage{multirow}
\usepackage{caption}

\begin{document}
%
% paper title
% Titles are generally capitalized except for words such as a, an, and, as,
% at, but, by, for, in, nor, of, on, or, the, to and up, which are usually
% not capitalized unless they are the first or last word of the title.
% Linebreaks \\ can be used within to get better formatting as desired.
% Do not put math or special symbols in the title.
\title{CUNet: A Compact Unsupervised Network for Image Classification}
%
%
% author names and IEEE memberships
% note positions of commas and nonbreaking spaces ( ~ ) LaTeX will not break
% a structure at a ~ so this keeps an author's name from being broken across
% two lines.
% use \thanks{} to gain access to the first footnote area
% a separate \thanks must be used for each paragraph as LaTeX2e's \thanks
% was not built to handle multiple paragraphs
%

\author{Le~Dong,~\IEEEmembership{Member,~IEEE,}
        Ling~He,~Gaipeng~Kong,
        ~Qianni~Zhang,~Xiaochun Cao,~and Ebroul Izquierdo% <-this % stops a space
\IEEEcompsocitemizethanks{\IEEEcompsocthanksitem L. Dong, L. He and G. Kong are with the School of  Computer Science and Engineering, University of Electronic Science and Technology of China (UESTC), 2006 Xiyuan Avenue, Gaoxin West Zone, Chengdu, Sichuan, 611731, China.\protect\\
% note need leading \protect in front of \\ to get a newline within \thanks as
% \\ is fragile and will error, could use \hfil\break instead.
E-mail: ledong@uestc.edu.cn
\IEEEcompsocthanksitem Xiaochun Cao is with the Institute of Information Engineering, CAS.
\IEEEcompsocthanksitem Q. Zhang and Ebroul Izquierdo are with the School of Electronic Engineering and Computer Science, Queen Mary, University of London.

}% <-this % stops an unwanted space
\thanks{}}
% note the % following the last \IEEEmembership and also \thanks -
% these prevent an unwanted space from occurring between the last author name
% and the end of the author line. i.e., if you had this:
%
% \author{....lastname \thanks{...} \thanks{...} }
%                     ^------------^------------^----Do not want these spaces!
%
% a space would be appended to the last name and could cause every name on that
% line to be shifted left slightly. This is one of those "LaTeX things". For
% instance, "\textbf{A} \textbf{B}" will typeset as "A B" not "AB". To get
% "AB" then you have to do: "\textbf{A}\textbf{B}"
% \thanks is no different in this regard, so shield the last } of each \thanks
% that ends a line with a % and do not let a space in before the next \thanks.
% Spaces after \IEEEmembership other than the last one are OK (and needed) as
% you are supposed to have spaces between the names. For what it is worth,
% this is a minor point as most people would not even notice if the said evil
% space somehow managed to creep in.

% The paper headers
\markboth{Journal of IEEE TRANSACTIONS on Multimedia}%
{Shell \MakeLowercase{\textit{et al.}}: Bare Demo of IEEEtran.cls for Computer Society Journals}
% The only time the second header will appear is for the odd numbered pages
% after the title page when using the twoside option.
%
% *** Note that you probably will NOT want to include the author's ***
% *** name in the headers of peer review papers.                   ***
% You can use \ifCLASSOPTIONpeerreview for conditional compilation here if
% you desire.

% If you want to put a publisher's ID mark on the page you can do it like
% this:
%\IEEEpubid{0000--0000/00\$00.00~\copyright~2015 IEEE}
% Remember, if you use this you must call \IEEEpubidadjcol in the second
% column for its text to clear the IEEEpubid mark.

% use for special paper notices
%\IEEEspecialpapernotice{(Invited Paper)}

% make the title area
\maketitle

% As a general rule, do not put math, special symbols or citations
% in the abstract or keywords.
\begin{abstract}
In this paper, we propose a compact network called CUNet (compact unsupervised network) to counter the image classification challenge. Different from the traditional convolutional neural networks learning filters by the time-consuming stochastic gradient descent, CUNet learns the filter bank from diverse image patches with the simple K-means, which significantly avoids the requirement of scarce labeled training images, reduces the training consumption, and maintains the high discriminative ability. Besides, we propose a new pooling method named weighted pooling considering the different weight values of adjacent neurons, which helps to improve the robustness to small image distortions. In the output layer, CUNet integrates the feature maps gained in the last hidden layer, and straightforwardly computes histograms in non-overlapped blocks. To reduce feature redundancy, we implement the max-pooling operation on adjacent blocks to select the most competitive features. Comprehensive experiments are conducted to demonstrate the state-of-the-art classification performances with CUNet on CIFAR-10, STL-10, MNIST and Caltech101 benchmark datasets.

\end{abstract}

% Note that keywords are not normally used for peerreview papers.

% For peer review papers, you can put extra information on the cover
% page as needed:
% \ifCLASSOPTIONpeerreview
% \begin{center} \bfseries EDICS Category: 3-BBND \end{center}
% \fi
%
% For peerreview papers, this IEEEtran command inserts a page break and
% creates the second title. It will be ignored for other modes.
\IEEEpeerreviewmaketitle

\section{Introduction}
% The very first letter is a 2 line initial drop letter followed
% by the rest of the first word in caps.
%
% form to use if the first word consists of a single letter:
% \IEEEPARstart{A}{demo} file is ....
%
% form to use if you need the single drop letter followed by
% normal text (unknown if ever used by the IEEE):
% \IEEEPARstart{A}{}demo file is ....
%
% Some journals put the first two words in caps:
% \IEEEPARstart{T}{his demo} file is ....
%
% Here we have the typical use of a "T" for an initial drop letter
% and "HIS" in caps to complete the first word.
\IEEEPARstart{I}{mage} classification has long been a challenging task in vision community, especially when the image amount and intra-class variability get continually increasing. Numerous efforts have been made to counter this significant challenge, among which the bag-of-features (BoF) model has shown desirable performance. BoF works by extracting local features (e.g. SIFT) from the images, vector quantizing them and then representing images as histograms over the visual words. Thus, in the BoF representation, the spatial layout is completely discarded. As an extension of BoF, SPM (spatial pyramid matching) takes account of the spatial information of images, and has improved the classification performance on relatively small classification benchmarks like Caltech101 and Caltech256. However, such model design fails to demonstrate alike performance on mid-scale datasets such as STL-10 and large-scale datasets CIFAR-10. Parallel processing based on distributed resources [1] seems to release the bottleneck that the increasing image scale meets, while the improvement on processing algorithms should be the most fundamental solution. This question raised considerable interest in the subject of mid-level features [2, 3], and feature learning in general [4, 5, 6].

In recent years, deep convolutional neural networks (CNN) [7] have demonstrated breakthrough accuracies for large-scale image classification, which stimulates a hurry of study on further improving CNN architectures [8, 9, 10]. Training with sufficiently large and diverse datasets, these improved CNN networks successfully obtain state-of-the-art performance on visual recognition tasks. The success of CNN is mainly attributed to their ability to learn rich mid-level image representations instead of hand-designed low-level features. Typically, the convolutional neural networks adopt a three-stage formulation: filter bank convolution, neuron activation, and pooling. Among the aforementioned three stages, the filter bank convolution plays a central role. To learn an effective filter bank at each convolution stage, a variety of methods have been proposed, such as restricted Boltzmann machines (RBM) [11, 12], regularized auto-encoders and their variations [11]. In general, previous CNN networks optimize the filter bank by utilizing stochastic gradient descent (SGD) method on large number of labeled images, which strictly relies on the expertise of parameter initiation and fine tuning. In addition, such filter learning procedure is rather time-consuming, especially on a common CPU. With the emergence of GPU computing [13] and the fast deep learning framework Caffe, conventional CNN networks still seem to be promising. However, such hardware-based techniques just cannot relieve the aforementioned restrictions from the source. Besides, traditional CNN is based on supervised learning, which means that the image label is strictly required. However, nowadays, along with the increasing image scale, image label becomes scarce.

\begin{figure*}[ht]
\begin{center}
%\fbox{
\scalebox{1.2}[1.2]{\includegraphics[width=0.8\linewidth]{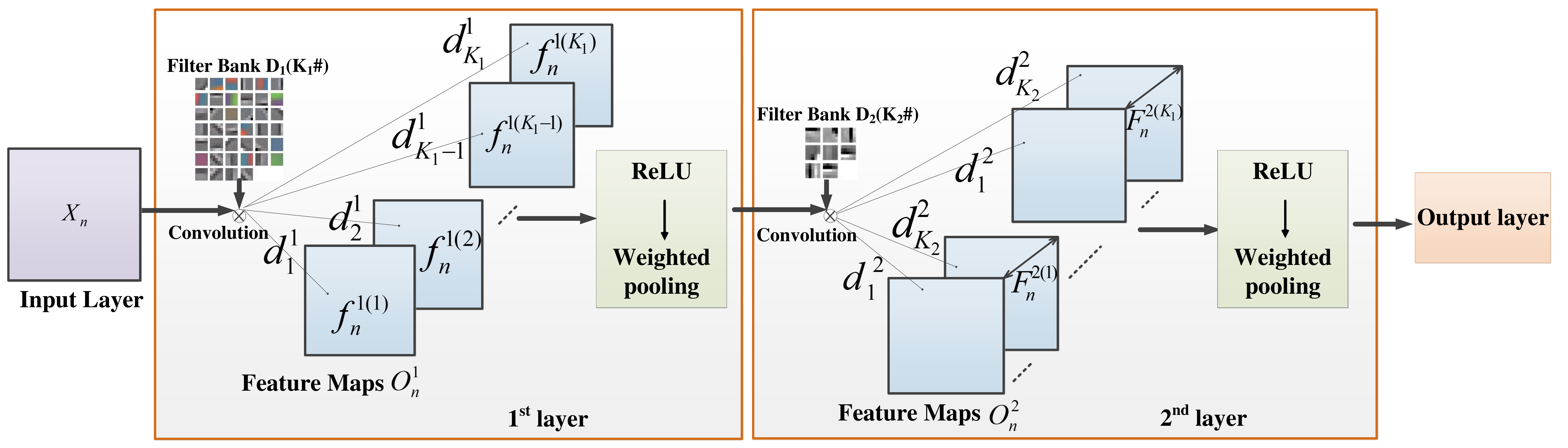}}
   %%The overview of our large-scale image retrieval system.
\end{center}
   \caption{CUNet structure.}
\label{fig1}
\end{figure*}

Considering that the success of current CNNs possesses a certain randomness due to the unsure filter bank learning procedure, researchers proposed another mathematically justified model named wavelet scattering networks (ScatNet) [14, 15, 16]. ScatNet is similar with CNN except for the design of its filter bank. The filter bank in ScatNet is simply predefined as wavelet operators, which significantly avoids the weights learning procedure. Despite the simpleness of the wavelet filter bank, [14] and [15] have verified that a similar multistage architecture of CNN leads ScatNet to accomplish superior performance on handwritten digit and texture recognition. However, such prefixed filter bank fails to capture information in diverse images, which makes it hard to be generalized to show competent performance in arbitrary vision tasks.

In this paper, we propose to construct a compact unsupervised network (CUNet) for image classification, which consists of the simpleness of filter bank in ScatNet and the generalized ability of CNN. Specifically, we straightforwardly use the classical K-means to learn the filter bank from randomly extracted image patches. Here, the scarce labeled images are not necessary, unlabeled ones are engouth to train filters. After the convolution, we maintains the Rectified Linear Units  (ReLUs) to activate neurons, followed by the proposed weighted pooling. Subsequent hidden layers are constructed in the same way, except that the filter banks are learned from previous feature map patches. In the output layer, each neuron is binary-mapped, and each group of feature maps are integrated to coarsely represent the input image. Then, histograms are straightforwardly computed in each non-overlapped block, followed by the max-pooling operation on the adjacent blocks to reduce the feature redundancy and select the most competitive features.

The contribution of our proposed CUNet can be concluded in three aspects:

\textbf{(1)  The filter bank learning procedure is compact and unsupervised, which abandons the millions of parameters initialization and fine tuning, and relieves the bottleneck of the scarce labeled images. Thus, CUNet effectively avoids falling into local optimum which traditional CNN usually suffers;}

\textbf{(2)  The proposed weighted pooling considers the different effects of all the activations in the pooling region, which contributes to improve the robustness to small image distortions;}

 \textbf{(3)  The histogram computing is a rather straightforward manner in image feature extraction. We choose to compute histograms in multiple blocks, which helps obtain the spatial information at a certain extent. The max-pooling trick further improve the feature competitiveness.}

The rest of the paper is organized as follows: Section 2 highlights the related works; Section 3 gives the formulation details of CUNet; Section 4 provides comprehensive experimental results to validate the superiority of CUNet; finally, Section 5 concludes the paper with directions for future work.

\section{Related Work}

Convolutional networks have recently demonstrated impressive progress in a variety of image classification and recognition tasks [13, 17, 18]. The promising perspective of CNNs stimulates researchers to make further study on this network for better performance. Multiple layers of unpooled convolution [7] have been utilized lately with considerable success, while such architectures must be carefully designed and sized using good intuition along with extensive trial-and-error experiments on a validation set. [19] proposed to transfer image representations learned with CNNs on large datasets to other visual recognition tasks with limited training data. Although [19] has achieved some success when reusing the ImageNet representation to compute mid-level image representation for the PASCALVOC dataset, it still needs tough training on ImageNet before the transfer operation. Besides, the representation learned from large datasets may incur overfitting issues when it is transferred to small datasets. [8] proposed a new activation function called maxout to avoid pitfalls such as failing to use many of a model's filters, which make it possible to train deeper networks. Compared with conventional convolutional layers which perform linear separation, the maxout network is more potent as it can separate concepts that lie within convex sets. However, maxout network imposes the prior that instances of a latent concept lie within a convex set in the input space, which does not necessarily hold. [9] proposed the NIN network composed of mlpconv layers which use multilayer perceptrons to convolve the input and a global average pooling layer as a replacement for the fully connected layers in conventional CNN. While mlpconv layers model the local patches better and global average pooling prevents overfitting globally, NIN still faces the difficulty of millions of parameters training and fine tuning. Training recurrent neural networks usually incurs the vanishing and the exploding gradient problems. [20] proposed a gradient norm clipping strategy to deal with the exploding gradients problem, and used a regularization term that forces the error signal not to vanish as it travels back in time to relieve the vanishing gradient restriction. Though some improvements on the gradient training have been achieved, [20] still fails to simplify the inherent complex of current neural networks.

Our proposed CUNet does not use any image transformations or other regularization such as dropout [21] or maxout [8], only involves preprocessing image patches, learning K-means filter bank, computing histograms and selecting the most competitive histogram bins. Thus, our simplifications do not entail a departure from current methods in terms of performance.

% needed in second column of first page if using \IEEEpubid
%\IEEEpubidadjcol

\section{Compact Unsupervised Network}
In this section, we present the detail formulation of our proposed CUNet.  A two-layer CUNet structure is illustrated in Fig.1 where the output layer is precisely highlighted in Fig.2. In the next subsections, we will elaborate each component of the block diagram in detail.

\subsection{The pre-processing of the input layer}
Suppose we are given $N$ input training images $\left\{ X_n\right\}_{n=1}^N$ of size $W\times H \times d $, where $d=1$ for gray images and $d=3$ for RGB ones. CUNet begins by extracting random patches from the training images $\left\{ X_n\right\}_{n=1}^N$. Each $w-by-h$ patch can be denoted as a vector in $\mathscr{R}^M$ of pixel intensity values, with $M=w\times h\times d$. Then, we can construct a dataset containing $T$  randomly extracted patches, $P=\left\{p_1,\cdots,p_t,\cdots,p_T\right\}$, where $p_t\in \mathscr{R}^M$ . Given this patch dataset, we apply some necessary pre-processing operations on $P$ to obtain better configuration.

It is common practice for vision tasks to perform some simple normalization steps before attempting to generate features from the input data. In this work, each patch $p_t$  is normalized by subtracting the mean and dividing by the standard deviation of its elements. After normalizing each input vector,  we apply the whitening operation [23] over the whole dataset $P$. [29] has discussed the superiority of whitened images over non-whitened. Then, we obtain the pre-processed input dataset $\bar{P}=\left\{\bar{p_1},\cdots,\bar{p_2},\cdots,\bar{p_T} \right\}$. Assuming that the number of filters in the first layer is $K_1$, we run K-means on $\bar{P}$ to get the filter bank denoted as $D_1=\left\{d_1,\cdots,d_{k_1},\cdots,d_{K_1}\right\} \in \mathscr{R}^{M\times K_1}$ where each centroid $d_{k_1}$ will act as a convolution filter in the subsequent convolution stage.

\subsection{The formulation of the hidden layer}

We maintain the typical processing stages of the traditional CNN, \emph{i.e.}, filter convolution, pooling, neuron activation. Next, we will elaborate each stage added with special design constructed in CUNet.

\textbf{Filter convolution:} Given the first layer's  convolution filter bank $D_1=\left\{d_1,\cdots,d_{K_1} \right\}$ , we convolve each training image $X_n$ with the $K_1$ filters:
\begin{equation}
O_n^1=X_n\otimes D_1, n=1,\cdots,N,
\end{equation}
where $O_n^1=\left\{f_n^{1(1)},\cdots,f_n^{1(k_1)},\cdots,f_n^{1(K_1)}\right\}$ is the first layer's feature map set of $X_n$, $f_n^{1(k_1)}$  is the feature map of $X_n$ convolved by the filter $d_{k_1}$, and $\otimes$ denotes the $2D$ convolution operation.

\textbf{Nonlinear activation:} Then, the neurons in the feature maps need to be activated through a pre-defined activation function. The Tangent function $f(x)=tanh(x)$ and Sigmoid function $f(x)=(1+e^{-x})^{-1}$ are commonly used in previous networks and have been proved to be effective. However, considering the training time, these saturating nonlinearities are much slower than the non-saturating nonlinearity $f(x)=max(0,x)$. Following [24], we call the neurons activated by this nonlinearity Rectified Linear Units (ReLUs). [13]
has verified that deep convolutional neural networks with ReLUs train several times faster than their equivalents with tanh units. Therefore, CUNet likewise adopts ReLUs to accomplish subsequent efficient processing. In fact, we have tried the Tangent function and Sigmoid function in our CUNet while find they are not competitive with the ReLUs.

\textbf{Weighted pooling:} To build robustness to small distortions, we set pooling layer after the activation layer just as most of the traditional ConvNet architectures did. Conventional pooling usually includes two popular choices, \emph{i.e.} max pooling and average pooling. Max pooling always captures the largest response values, which may loose the useful information of the small ones. As for average pooling, it aggregates local statistics information by preventing large response values taking over and small ones being removed out. However, since average pooling treats each neuron equally, the usefulness of each neuron's response may be confused. [22] proposed the stochastic pooling, which replace the conventional deterministic pooling operations with a stochastic procedure, randomly picking the activation within each pooling region according to a multinomial distribution, given by the activities within the pooling region. Obviously, the choice of the multinomial distribution has dominated effect on the pooling performance. Inspired by these previous pooling strategy, we propose a new pooling method denominated weighted pooling, which considers each neuron's response as well as its response's usefulness account in the whole neurons' responses, that is, each neuron in the pooling region owns its proper weight. Suppose that the pooling window is of size $p_w-by-p_h$, the response value of each neuron is $a_{i,j}$ with $i=1,\cdots,w; j=1,\cdots,h$. Then, the pooling results of the $p_w-by-p_h$ window can be calculated according to Eq.(2):
\begin{equation}
P_{result}= w_{i,j} * a_{i,j}
\end{equation}
where $w_{i,j}$ is the weight of $a_{i,j}$. In this paper, we simply compute each neuron's value proportion in the pooling region as its weight, i.e., $ w_{i,j}=\frac{a_{i,j}}{\sum_{i}\sum_{j}a_{i,j}} $. The proposed weighted pooling will capture different proportion of local information of each neuron in the original feature map, thus leading to a better local representation. To validate the effectiveness of our proposed weighted pooling, we conducted experiments in Section 4 to compare the performance under different pooling strategies. Conventionally, pooling operation commonly summarizes the non-overlapping neighborhoods containing adjacent units, which reduces the computing complexity while leads to coarse pooling results. To be more precise, CUNet applies an overlapping sliding window with a stride $s$ to accomplish further accurate pooling results.

Thus, one layer of CUNet is accomplished, including three main stages: convolution, non-linear rectification, and weighted pooling. Note that the three steps maintain the feature map size as the original input image, the convolution and pooling operations both pad the images (or feature maps) with zeros. We tried the feature map size-maintained model and observe that it outperforms traditional size-changed ones. The second layer is in a similar formulation with the first layer, except that the filter bank $D_2$  is obtained by running K-means on the patches randomly extracted from the first layer's output. Certainly, we can stack multiple layers as previous works [9] to gain higher level features, whereas, we find that two layers beyond provide subtle performance improvement. Thus, our CUNet adopts a two-layer architecture, and a deeper model can be implemented, where applicable, in the same way.

\subsection{The design of the output layer}
The detailed design of the output layer is highlighted in Fig.2. In the second layer, each of the $K_1$ feature maps $f_n^{1(k_1)}$ has $K_2$  outputs $F_n^{2(k_1)}=\left\{f_n^{2(1)},f_n^{2(2)},\cdots,f_n^{2(K_2)}\right\}$ . First, each set of the $K_2$ feature maps are binary mapped, where each unit value is set as 1 if it is positive and 0 if non-positive. Thus, the feature maps are all composed of ones and zeros, and we call these B-maps. Obviously, such crude mapping inevitably loses some useful feature information. To obtain complementary feature information and inspired by [25], we integrate the $K_2$ B-maps in $F_n^{2(k_1)}$ into one integer-valued image with each feature map multiplying a coefficient $\lambda_i$:
\begin{equation}
I=\sum_{i=1}^{K_2}\lambda_if_n^{2(k_2)},
\end{equation}
where $\lambda_i=2^{i-1}$, $f_n^{2(k_2)}$ is the $k_2$-th B-map in $F_n^{2(k_1)}$. The order and weights of the $K_2$  B-maps does not have any relevant effects on the network performance.

\begin{figure}[ht]
\begin{center}
%\fbox{
\scalebox{1.2}[1.2]{\includegraphics[width=0.8\linewidth]{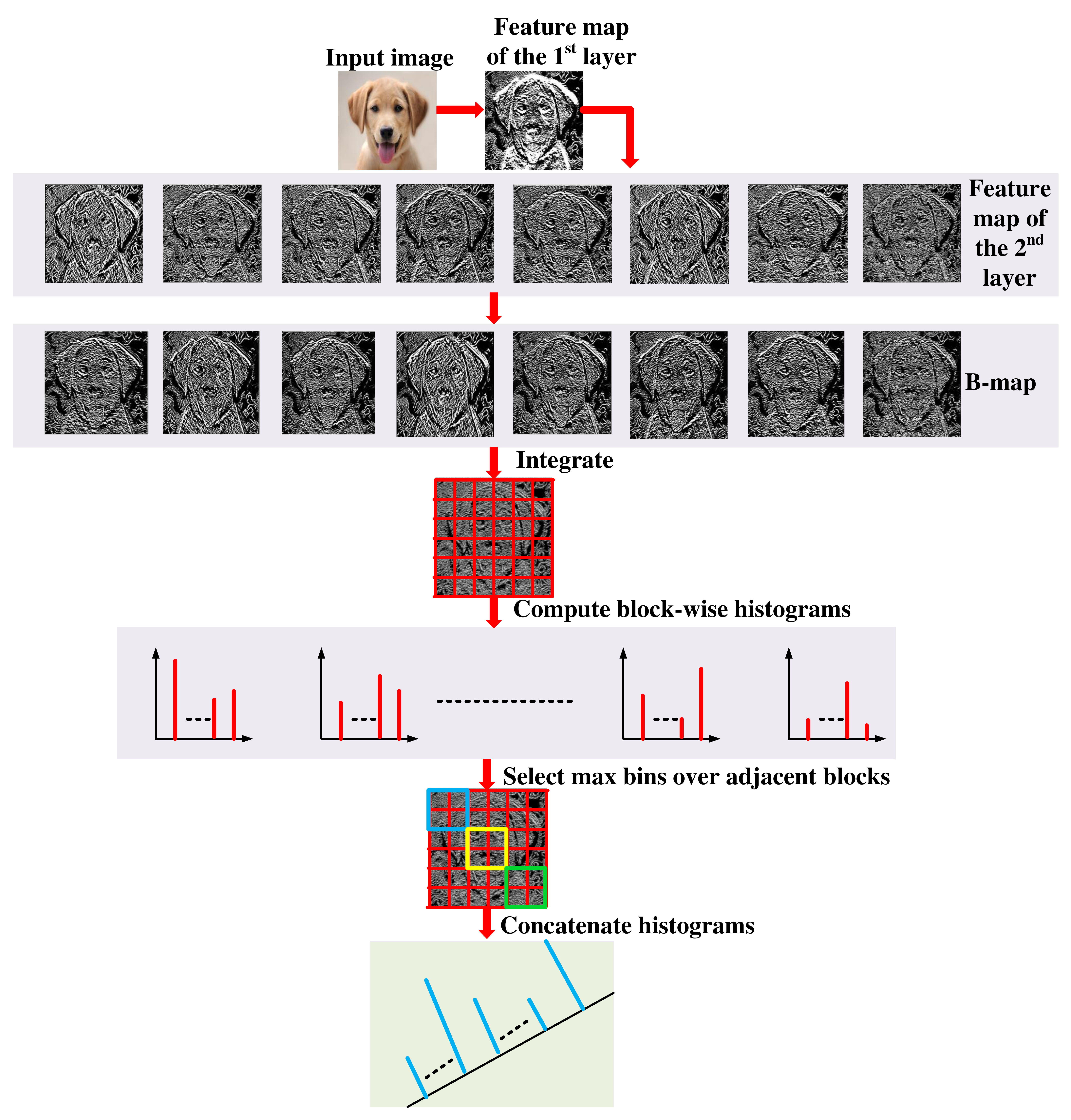}}
   %%The overview of our large-scale image retrieval system.
\end{center}
   \caption{Details of CUNet output layer.}
\label{fig2}
\end{figure}

For each of the $K_1$ feature maps in $O_n^1$ , we can obtain its corresponding image $I_{k_1}$ with $k_1=1,\cdots,K_1$. Next, we simply compute the histogram of each $I_{k_1}$  to gain the final image representation. Considering the robustness that geometric invariance brings to image classification and matching of highly variable scenes, we propose to compute the histogram in a window-wise manner. [25] similarly adopts such strategy and achieves desirable performance. However, we argue that such histogram computing incurs feature redundancy and high dimension problem. In order to release this restriction, we implement the max-pooling operation on histogram bins in adjacent blocks. In particular, for the adjacent $w*w$ blocks,  we select the max bins in each block. Thus, these $w*w$ histograms results in one histogram. Such max-pooling operation helps obtain the most competitive image feature, avoids feature redundancy, and controls the feature dimension in a reasonable scope. Finally, we concatenate the histograms gained from each group of $w*w$ blocks as the image feature, followed by a liblinear SVM as classifier to classify the images.

\section{Experimental Evaluation}

We evaluate the performance of CUNet on four benchmark datasets: STL-10, Caltech101, CIFAR-10, and MNIST. The networks used for the four datasets all consist of two stacked layers, followed by a linear SVM classifier. More particular experimental settings are presented in subsequent sections. we quote results directly from the literature to give a comparison since we note that sometimes we could not reproduce previous works¡¯ results, largely due to subtle engineering details.

\subsection{The Classification Performance}

\subsubsection{CIFAR-10}

The CIFAR-10 dataset is composed of 10 classes of natural images split into 50,000 for training and 10,000 for testing. Each image is an RGB one of size 32-by-32. Images vary greatly within each class not only in object position and object size, but also in colors and textures. Besides, the background of each image shows significant variance.

In particular, we learn $K_1= 40$ filters of size $5 \times 5$ in the first layer and $K_2= 8$ of size $5 \times 5$ in the second layer. Both the two layers set their weighted pooling size as 2$\times$2, and the pooling windows are overlapped with one pixel stride. The histogram computing blocks are all of size $4*4$, non-overlapped. After getting the block-wise histograms, we select the max bins over adjacent $2*2 = 4$ blocks into one single histogram.

TABLE 1 presents the classification accuracy of different methods on CIFAR-10. We observe that CUNet, with weighted pooling, achieves desirable performance among these methods. Besides, the results show that the pooling strategy somewhat influences the final classification performance when all the other settings keep the same. Among the three pooling strategies (\emph{i.e.}, our proposed weighted pooling, the prevalent max and average pooling), our weighted pooling shows better performance, about 0.38\% higher than max pooling and 0.85\% higher than average pooling. Note that the filter banks used in the condition of weighted pooling are maintained to work in max and average pooling, which strictly avoids the subtle influence of filters randomly learned by K-means. This rigid experimental setting is similarly applied to STL-10, Caltech101, and MNIST for a fair condition.

\begin{table}[h]
\centering \caption{Comparison of accuracy(\%) of different methods on CIFAR-10 with no data augmentation.}
\begin{tabular}{c|c}
\hline Methods & Accuracy(\%)\\
\hline CUNet $+$  Weighted pooling &80.31\\
\hline CUNet $+$ Max pooling &79.93\\
\hline CUNet $+$ Average pooling &79.46\\
\hline Tiled CNN [26]    &73.10\\
\hline Improved LCC [27]    &74.50\\
\hline KDES-A [28]      &76.00\\
\hline K-means (Triangle,4000features) [29]      &79.60\\
\hline Cuda-convnet2 [30]      &82.00\\
\hline CKN-CO [10]         &82.18\\
\hline Discriminative SPN [31]         &83.96\\
\hline TIOMP-1/T (combined, K= 4,000) [32]         &82.20\\
\hline 2x PDL (1600 codes) [33]         &78.71\\ \hline
\end{tabular}
\end{table}

\subsubsection{STL-10}

The STL-10 dataset consists of 96-by-96 pixels color images belonging to 10 different classes. This dataset is inspired by the CIFAR-10 while providing fewer training examples (500 per class) and test examples (800 per class), which forces algorithms to rely heavily on acquired prior knowledge of image statistics. We downsampled the STL-10 images into $32 \times 32$ pixels for a simpler configuration.

Experimental settings for STL-10 are similar with CIFAR-10, except that $K_1= 30$ filters.
TABLE 2 gives the comparison of different methods on STL-10. We observe that CUNet, with weighted pooling, provides desirable performance among these previous works. With the other settings keeping the same, weighted pooling helps increase the classification accuracy by 0.6\% (max pooling) and 0.4\%(average pooling).

\begin{table}[h]
\centering \caption{Comparison of accuracy(\%) of different methods on STL-10 with no data augmentation.}
\begin{tabular}{c|c}
\hline Methods & Accuracy(\%)\\
\hline CUNet $+$ Weighted pooling &63.00\\
\hline CUNet $+$ Max pooling &62.40\\
\hline CUNet $+$ Average pooling &62.60\\
\hline 2x PDL (1600 codes) [33]         &58.28\\
\hline CKN-CO [10]         &62.32\\
\hline EPLS [34]         &61.00\\
\hline Discriminative SPN [31]         &62.30\\
\hline sparse TIRBM (combined) [32]         &58.70\\
\hline
\end{tabular}
\end{table}

We list some sample images from each class in Fig.3, and the classification accuracy of each class is labeled next to the corresponding image rows. From the results, we observe that the relatively simple classes commonly achieve higher accuracy, such as airplane(81.38\%), ship(81.00\%), and car(80.13\%). These aforementioned three objects are all present relatively simplex appearance. Besides, they are static objects and thus do not incur the confusing problems such as activity variance. Differently, the living objects such as monkey(53.50\%), cat(43.50\%), and dog(31.00\%) commonly gives lower accuracy. From the sample images, we observe that such animals commonly includes various kinds, and they usually show different actions, even hidden by some other obstruction, which undoubtedly brings classification difficulty.

\begin{figure}[ht]
\begin{center}
%\fbox{
\scalebox{0.8}[0.8]{\includegraphics[width=0.8\linewidth]{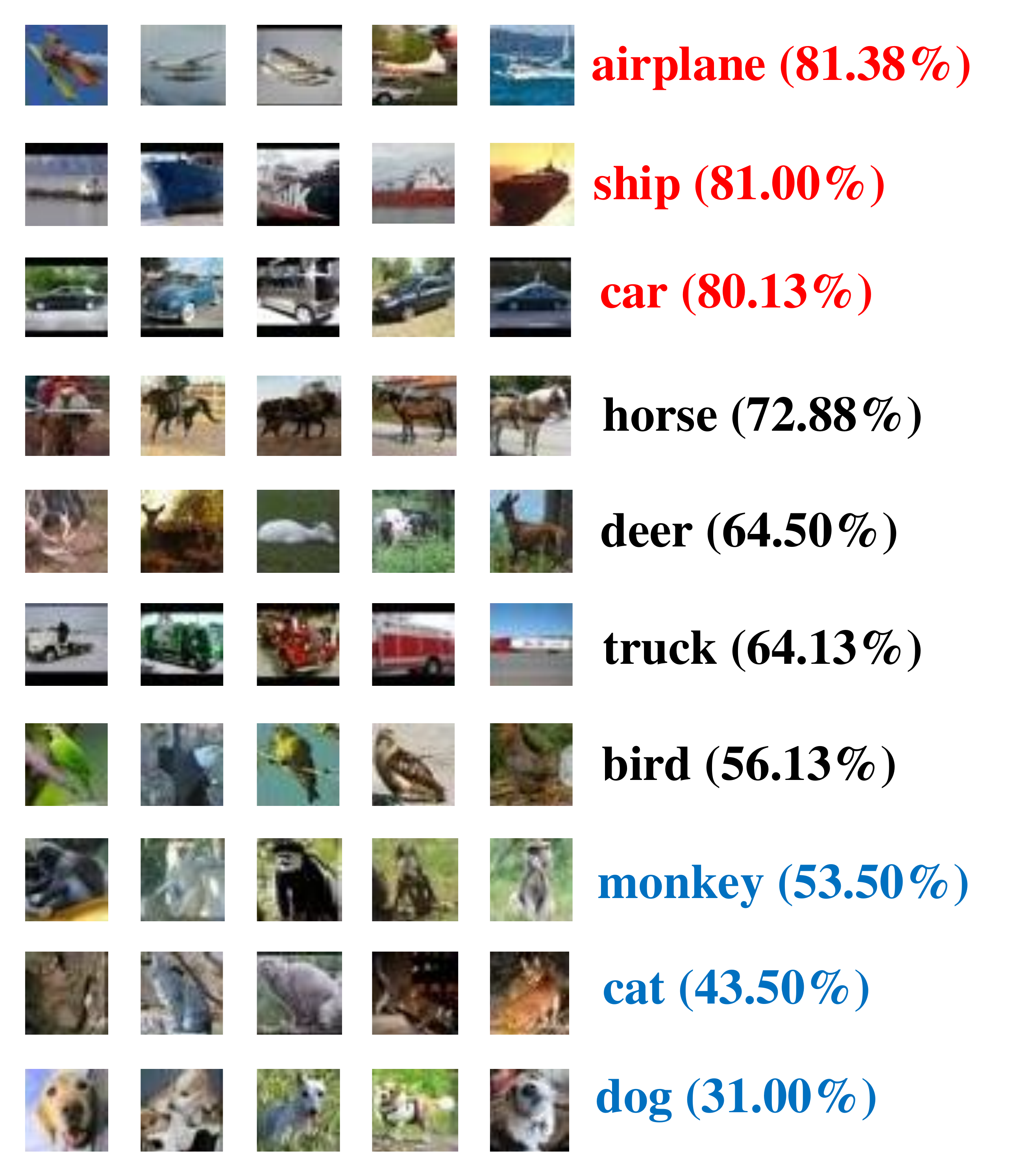}}
   %%The overview of our large-scale image retrieval system.
\end{center}
   \caption{Classification accuracy of each class in STL-10.}
\label{fig3}
\end{figure}

\subsubsection{Caltech101}

Caltech101 dataset contains 101 classes (including animals, vehicles, flowers, \emph{etc.}) with significant variance in shape, and a background class. The number of images per category varies from 31 to 800. For experimental convenience, we convert all the images into grey, and resize the images into $32*32$ without keeping the aspect ratio. Following the traditional settings, we randomly select 15 and 30 train images per class (including the background class), respectively. TABLE 3 presents the classification results on Caltech101. For both 15 and 30 training images (per class), we train $K_1 = 30$ filters. Other settings are the same with CIFAR-10.

From TABLE 3, we observe that the proposed CUNet with weighted pooling achieves desirable performance among current state-of-the-arts methods based on raw pixels. Note that we directly resize the images into $32*32$ without keeping the aspect ratio, while previous methods commonly adopts some strategies to maintain the aspect ratio of the images. Even though, our proposed CUNet still shows its competitiveness on Caltech101. Similar with CIFAR-10 and STTL-10, the weighted pooling successfully outperforms max pooling and average pooling at different extent.

\begin{table}[h]
\centering \caption{Comparison of accuracy(\%) of different methods on Caltech101}
\begin{tabular}{c|c|c}
\hline Training size & 15 & 30\\
\hline
\hline CUNet $+$  Weighted pooling      & 58.62   &  66.72 \\
\hline CUNet $+$  Max pooling         &58.00      &  66.34\\
\hline CUNet $+$  Average pooling      &58.14     &  66.48\\
\hline CDBN [38]          & 57.70 & 65.40 \\
\hline ConvNet [39]      & 57.60 & 66.30\\
\hline DeconvNet [40]    & 58.60 & 66.90\\
\hline Chen \emph{et al.} [36]     & 58.20 & 65.80\\
\hline Zou \emph{et al.} [37]     & - & 66.50\\
\hline
\end{tabular}
\end{table}

\begin{figure*}[ht]
\begin{center}
%\fbox{
\scalebox{1.2}[1.2]{\includegraphics[width=0.8\linewidth]{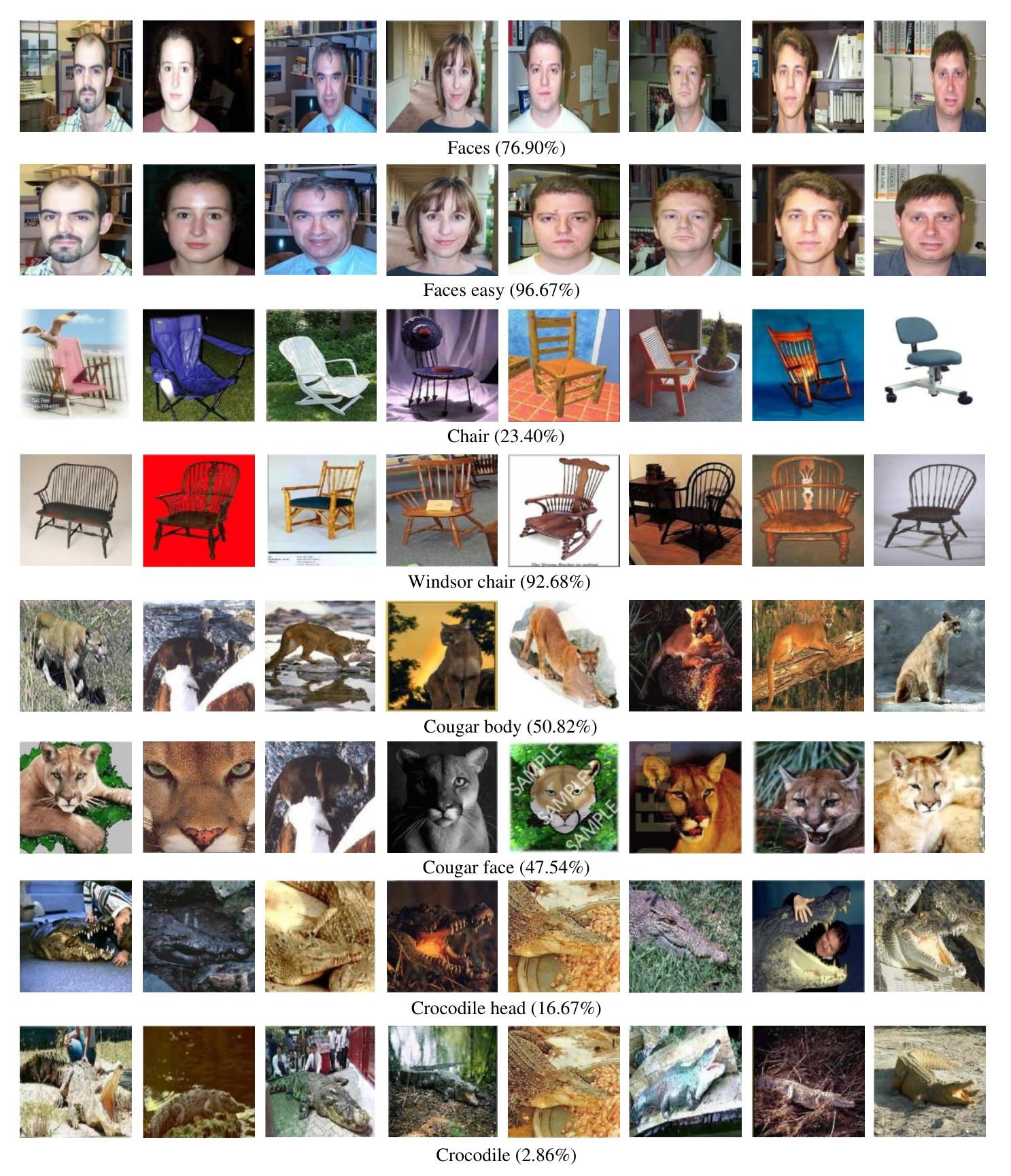}}
   %%The overview of our large-scale image retrieval system.
\end{center}
   \caption{Caltech101 results(15 train images per class).}
\label{fig4}
\end{figure*}

Fig.4 shows some classification results of Caltech101. Each two contiguous rows are two classes that have little inter-class difference. From the result, we observe that the classification accuracy between each two rows shows a big gap. For example, the classification accuracy of the class \textbf{Faces} is 76.90\%, about 19.77\% lower than the class \textbf{Faces easy}(96.67\%). From the listed example images, we observe that the faces in the class \textbf{Faces easy} are exactly in the center of the images and little background are included, while the positions of faces in the class \textbf{Faces} are random (left or right, but no center), and all the images present a complex background, which makes the main object (face) become confusing. Therefore, \textbf{Faces} is more difficult to classify than \textbf{Faces easy}. Besides, once the
images are roughly resized, the objects of \textbf{Faces} are commonly get distorted, which further
brings classification difficulty to \textbf{Faces}. The two class \textbf{Chair} and \textbf{Windsor chair}
similarly show great difference on classification performance. \textbf{Chair} (23.40\%) gives 69.28\%
lower accuracy than \textbf{Windsor chair} (92.68\%). From the listed example images,
it is obvious that the object in \textbf{Chair} varies greatly and the background is relatively complex.
Differently, the intra-class variability of \textbf{Windsor chair} is subtle, and the background is much
more simpler than \textbf{Chair}. Thus, it is not surprising why the \textbf{Windsor chair} classification
performance is much higher than \textbf{Chair}. Similar analysis goes for the listed \textbf{Cougar body} (50.82\%)
and \textbf{Cougar face} (47.54\%), \textbf{Crocodile head} (16.67\%) and \textbf{Crocodile} (2.86\%).
From aforementioned discussion, we argue that CUNet is competitive in classifying the classes that show simple background,
little intra-class variability, and obvious object. However, we have to admit that CUNet shows less competitiveness in the
classes that have complex background, great intra-class variability, and confusing object.

\subsubsection{MNIST}

The basic MNIST dataset consists of 28-by-28 greyscale images of handwritten digits 0-9, with 10,000 training, 2,000 validation, and 10,000 test examples. To conveniently obey the processing baseline, we resize each MNIST image into $32*32$, and keep other settings the same with aforementioned three datasets, except that the filter number $K_1 = 5$.

TABLE 4 gives the classification error rate on basic MNIST with different methods. Still, the proposed weighted pooling outperforms the average and max pooling. Since MNIST is a relatively simple dataset, all methods perform well and close, thus, the subtle performance difference is not so statistically meaningful.

\begin{table}[h]
\centering \caption{Comparison of error rate(\%) of different methods on MNIST with no data augmentation.}
\begin{tabular}{c|c}
\hline Methods & Error rate(\%)\\
\hline CUNet $+$ Weighted pooling  &1.80\\
\hline CUNet $+$ Max pooling  &1.86\\
\hline CUNet $+$ Average pooling  &1.90\\
\hline CAE-2 [35]         &2.48\\
\hline ScatNet-2 [14]         &1.27\\
\hline
\end{tabular}
\end{table}

Fig.4 presents some MNIST training examples and the corresponding error rate. From Fig.4, we observe that both the simple classes \textbf{0} and \textbf{1} show little intra-class variance, and these two classes achieve lowest error rate. As for the other more complex digits which show great intra-class variance, they achieve higher error rate at different extent. From the sample images, we find it is even hard to artificially judge the these confusing digits.

\begin{figure}[ht]
\begin{center}
%\fbox{
\scalebox{0.8}[0.8]{\includegraphics[width=0.8\linewidth]{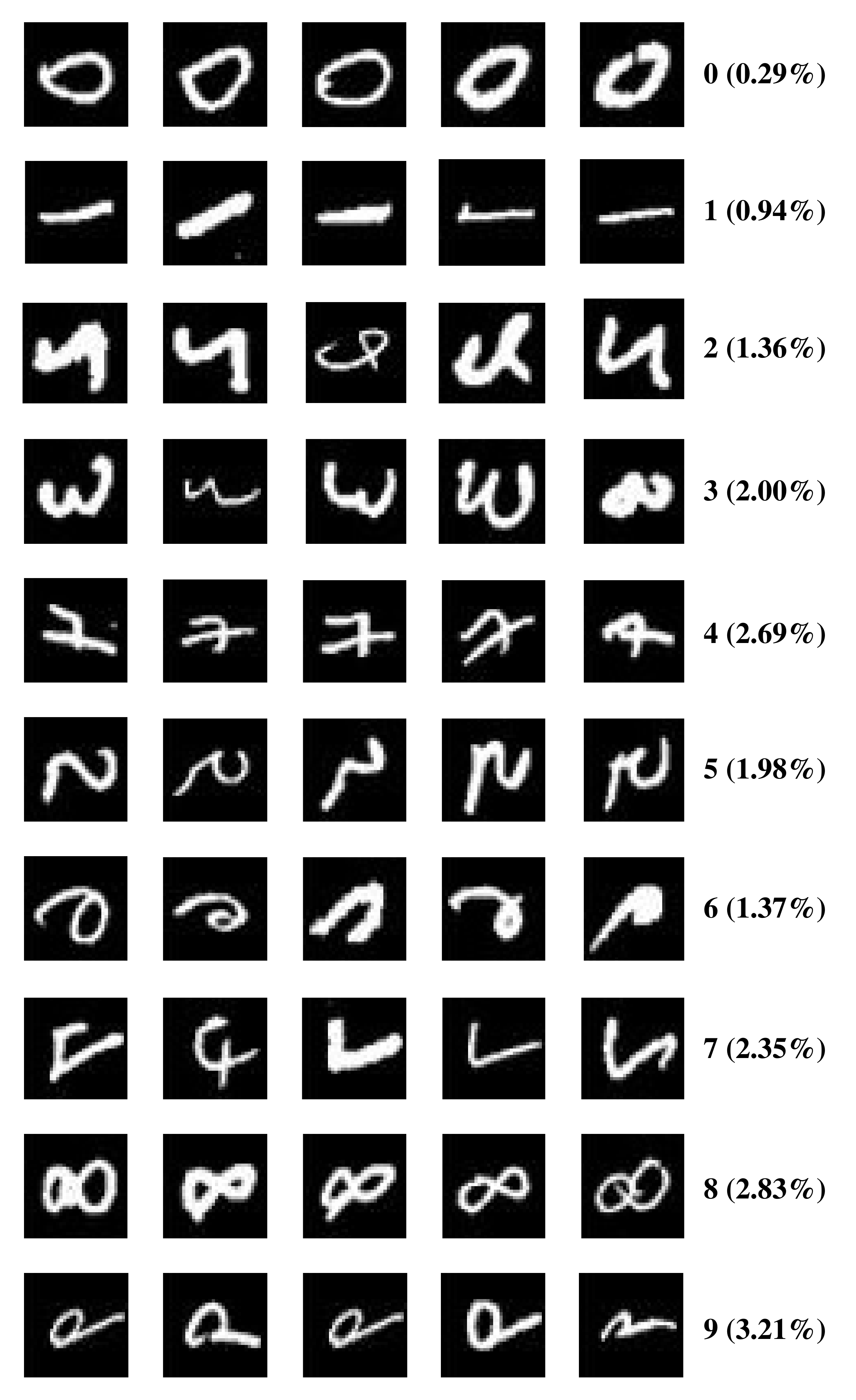}}
   %%The overview of our large-scale image retrieval system.
\end{center}
   \caption{Classification error rate of each class in basic MNIST.}
\label{fig5}
\end{figure}

\subsection{Impact of the number of filters}

In this section, we conducted experiments to validate the impact of the filter number on CUNet performance. We fix the experimental settings as aforementioned in Section 4.1 (\emph{i.e.}, the settings make each dataset gain its best performance), only change the filter number of the first layer. In particular, since CIFAR-10 is a relatively complicated dataset, we vary $K_1$ from 20 to 40. For the mid-scale STL-10 and Caltech101, we vary $K_1$ from 10 to 30. For the simpler MNIST, we vary $K_1$ from 5 to 15.

\begin{figure}[ht]
\begin{center}
%\fbox{
\scalebox{1.0}[1.0]{\includegraphics[width=0.8\linewidth]{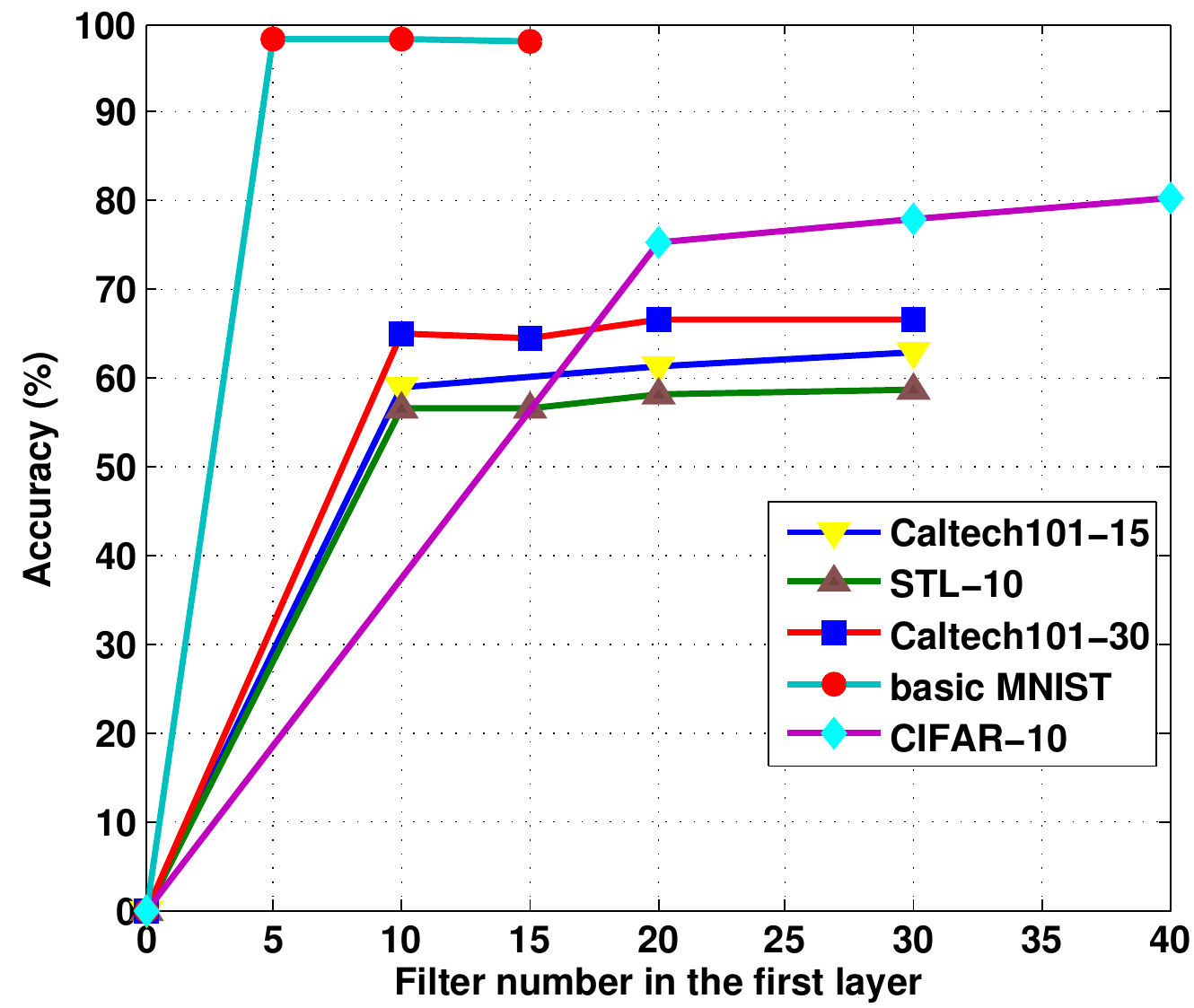}}
   %%The overview of our large-scale image retrieval system.
\end{center}
   \caption{Impact of the filter number on classification accuracy.}
\label{fig6}
\end{figure}

Fig.6 illustrates the impact of the filter number on classification performance. From Fig.6, we observe that the classification accuracy will get improvement when the filter number increases. Whatever the dataset is, more filters will certainly help improve the classification performance as we expect. However, such increase is not always existent. When the filter number achieves its saturation value, the classification performance shows subtle improvement. We attribute such phenomenon to the useless duplicates of the filters, that is, some filters will be repeated if the filter number is set larger than the saturation value. The repeated filters will contribute nothing and even drag down the final classification performance. Hence, the set of filter number plays some dominated role in CUNet.

\subsection{Impact of the block size}

Fig.7 illustrates the impact of the block size on CUNet performance. Here, the block size refers to the width and height of the histogram computing windows. For each of the dataset, we set the block size as $4*4$, $8*8$, $16*16$, and fix other settings as discussed in Section 4.1. From Fig.7, we observe that the classification performance get decreased when the block size increases. Commonly, for all the datasets, the classification accuracy achieves the highest when the block size is $4*4$. When the block size increases to $8*8$, the classification performance get undesirable decrease, and such performance decline is even larger when the block size increases to $16*16$. However, it is deserved to be mentioned that although the classification accuracy goes down along with the increase of block size, feature dimension also gets decreased, which obviously brings computation release to the experimental devices. In particular, when the block size is $4*4$, the feature dimension is $K_1 * 2^{K_{2}} * 16$, while when the block size is set as $8*8$, the feature dimension is $K_1 * 2^{K_{2}} * 4$, 4 times dimension decrease. When the block size is $16*16$, the feature dimension is $K_1 * 2^{K_{2}} * 1$, 16 times decrease compared with the feature dimension of block size $4*4$.

\begin{figure}[ht]
\begin{center}
%\fbox{
\scalebox{1.0}[1.0]{\includegraphics[width=0.8\linewidth]{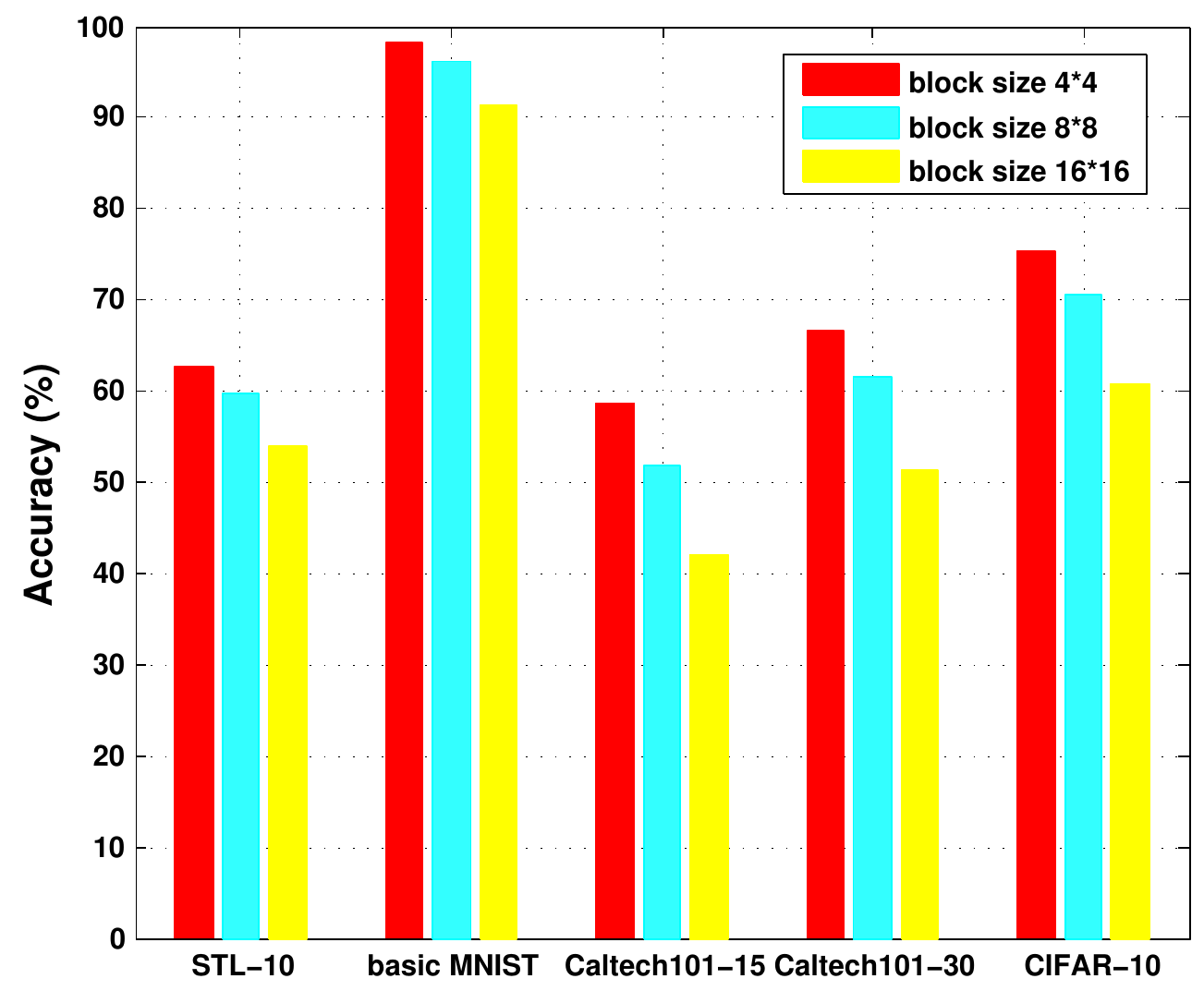}}
   %%The overview of our large-scale image retrieval system.
\end{center}
   \caption{Impact of the block size on classification accuracy.}
\label{fig7}
\end{figure}

Based on aforementioned analysis, the block size has two-way influence on CUNet. On the one hand, the increase of block size results in performance decline. On the other hand, feature dimension desirably decreases along with the increase of the block size. Hence, the choice of the block size is largely depended on one's own focus. If accuracy is strictly required, then the block size should set smaller. On the contrary, if the experimental devices can not meet the dataset scale, then larger block size should be set.

\subsection{Discussion}

Aforementioned experiments have successfully validated the effectiveness of CUNet from different aspects.  Four datasets (CIFAR-10, STL-10, Caltech101, MNIST) are
employed to test the performance of CUNet on different classification
tasks. Firstly, we provide image classification accuracy on these four
datasets to validate the feasibility of CUNet. The accuracies of some example classes are presented to analyze the superiority and inferiority of CUNet. From the result, we found that CUNet is more competitive on those static objects (\emph{e.g.} , airplane, ship, car) and those showing little inner-class variability. Correspondingly, CUNet presents less competitive ability on those dynamic objects (\emph{e.g. }, dog, cat, monkey) and those showing little intra-class variability.
Secondly, we test the effect of the inner settings of CUNet on
classification performance: 1) we found that whatever the dataset is, more
filters will certainly help improve the classification performance, but
such increase is not always existent. When the filter number achieves its
saturation value, the classification performance shows subtle
improvement ; 2) the choice of the block size largely depends on real applications. The classification accuracy goes down along with the increase of block size, however, feature dimension also gets decreased, which obviously brings computation release to the experimental devices.

\section{Conclusion}

We propose a compact unsupervised network called CUNet for image classification tasks. The main purpose of the proposed CUNet is to simplify the traditional convolutional neural network. CUNet is compact which avoids millions of parameters tuning and does not require numerical optimization solver. Besides, unsupervisedly learning convolution filters addresses the bottleneck of scarce image labels. Experimental results verify that CUNet is competitive among previous state-of-the-art works. In future work, we would like to further simplify CUNet and make it feasible for more challenging large-scale dataset benchmarks, especially those own great intra-class variance.

% if have a single appendix:
%\appendix[Proof of the Zonklar Equations]
% or
%\appendix  % for no appendix heading
% do not use \section anymore after \appendix, only \section*
% is possibly needed

% use appendices with more than one appendix
% then use \section to start each appendix
% you must declare a \section before using any
% \subsection or using \label (\appendices by itself
% starts a section numbered zero.)
%

%\appendices
%\section{Proof of the First Zonklar Equation}
%Appendix one text goes here.
%
%% you can choose not to have a title for an appendix
%% if you want by leaving the argument blank
%\section{}
%Appendix two text goes here.

% use section* for acknowledgment
%\section*{Acknowledgment}
%
%
%The authors would like to thank...

% use section* for acknowledgment
\ifCLASSOPTIONcompsoc
  % The Computer Society usually uses the plural form
  \section*{Acknowledgments}
\else
  % regular IEEE prefers the singular form
  \section*{Acknowledgment}
\fi

This work was supported in part by the National Natural Science Foundation of China
under Grant 61370149, in part by the Fundamental Research Funds for the Central
Universities (No. ZYGX2013J083), and in part by the Scientific Research Foundation
for the Returned Overseas Chinese Scholars, State Education Ministry.

% Can use something like this to put references on a page
% by themselves when using endfloat and the captionsoff option.
\ifCLASSOPTIONcaptionsoff
  \newpage
\fi

\end{document}